# Curation of a Cardiology Interface Terminology for Highlighting Electronic Health Records using Machine Learning

Mahshad Koohi Habibi Dehkordi[1], Shuxin Zhou[2], Yehoshua Perl[3], Fadi P. Deek[4], James Geller[5], Gai Elhanan[6], Andrew J. Einstein[7], Luke Lindemann[8], Vipina K. Keloth[9]

[1] Department of Computer Science, New Jersey Institute of Technology, Newark, NJ, USA, mk985@njit.edu
[2] Department of Computer Science, St.Francis College, Brooklyn, NY, USA, szhou@sfc.edu
[3] Department of Computer Science, New Jersey Institute of Technology, Newark, NJ, USA, yehoshua.perl@njit.edu
[4] Department of Informatics, New Jersey Institute of Technology, Newark, NJ, USA, fadi.deek@njit.edu
[5] Department of Data Science, New Jersey Institute of Technology, Newark, NJ, USA, james.geller@njit.edu
[6] Center for Genomic Medicine, School of Medicine, University of Nevada, Reno, NV, USA, gelhanan@unr.edu
[7] Department of Medicine, Cardiology Division, Columbia University Irving Medical Center, NY, USA, ae2214@cumc.columbia.edu
[8] Advanced Metrics Laboratory, School of Medicine and Health Sciences, George Washington University, Washington, DC, USA, lukelindemann@gwmail.gwu.edu
[9] Department of Biomedical Informatics and Data Science, Yale University, New Haven, CT, USA, vipina.kuttichikeloth@yale.edu

**Corresponding author:** Mahshad Koohi Habibi Dehkordi, Department of Computer Science, New Jersey Institute of Technology, Ying Wu College of Computing, Suite 4100, University Heights, Newark, New Jersey 07102, USA. Email: mk985@njit.edu

# Abstract

Electronic health record (EHR) notes are dense medical documents containing large amounts of information, often filled with complex medical jargon. Highlighting all details in EHRs helps reduce the likelihood of missing crucial information by drawing attention to key content. This study proposes the design of a Cardiology Interface Terminology (CIT) to accurately highlight all details in EHR notes of cardiology patients.

We introduce an innovative Machine Learning (ML) technique for the design of CIT. The ML technique requires training data. Manual preparation of such training data is time-consuming and expensive. The process of the CIT design includes three phases. In the first two phases, we innovatively derive a training data CIT to be used by the third phase, ML technique.

We start by designing an initial CIT, composed of several components: the cardiology-related sub-hierarchies of SNOMED, other SNOMED concepts mined from EHRs of build set, and necessary components of terms e.g., medical abbreviations and medications. Utilizing an iterative process, fine-grained phrases containing initial CIT concepts are extracted from build set as CIT concept candidates. The candidate concepts are semi-automatically reviewed before being added to CIT, yielding the training data CIT, TCIT.

In the third phase, a ML model is trained with TCIT to identify candidates fitting to be concepts in the CIT. This model is used to extract further concepts from build set, yielding the final CIT.

The final CIT is then used to highlight the test set and evaluate the extent to which it captures details in an unseen EHR dataset. For this purpose, four evaluation metrics, coverage, breadth, completeness, and conciseness are used.

The highlighted test set has a coverage of 74.21%, with a breadth of 1.68. For 20 random notes in test set, the average completeness is 98.2% and average conciseness is 84.2%.

# 1 Introduction

Electronic health records (EHR) notes are dense medical documentation containing vast amounts of patient information, often written with abbreviations and complex medical jargons [1-3]. Highlighting EHRs involves marking all occurrences of concepts from a given terminology within the text. The highlighted text is intended to mark all the details in the text, which is different from “annotation.” Annotation is an example of highlighting, which is limited to marking only the concepts of a reference terminology and identifying the type of information, e.g., medications and treatments [4]. While in highlighting an EHR note we try to capture all the details, in annotation we typically are limiting ourselves to capture only some desired details according to the given research.

Due to the demanding nature of their work, medical professionals are often short on time. The need to quickly review many EHRs [5, 6] compels some to resort to skimming the clinical notes [7], which increases the risk of missing important details. When a clinician skims only the highlighted text, time is saved by focusing on the important content. Reading only the highlighted portion of a larger text, including the parsing into granular concepts, also reduces the implicit mental effort required to select which parts to read. Summarizing and simplifying EHRs is important because discharge notes and other clinical documentation are often long, filled with medical abbreviations, and difficult to understand for both patients. Clinicians may overlook details when reviewing lengthy notes, and patients frequently struggle to interpret medical terminology, abbreviations, and dense descriptions of their conditions or treatments. Accurate summarization can help clinicians quickly grasp essential information, while simplification supports patient comprehension, facilitates shared clinician-patient decision-making, and can improve adherence to treatment plans.

Large Language Models (LLMs) [8, 9] such as GPT-4o [10] have shown promise in summarizing and simplifying EHRs [8, 11, 12]. However, notes generated by LLMs can also introduce inaccuracies known as hallucination. In [13, 14], we demonstrated that summarizing highlighted EHRs with LLMs yields summaries that are more accurate and include more details compared to the summaries of unhighlighted versions. This is because the highlighted text helps focus on the details of the clinical note. Hence, highlighting EHRs can improve the accuracy of their LLM summaries. This approach has practical implications, particularly in simplifying EHRs for patients without sufficient medical knowledge.

In addition, effective highlighting of cardiology EHR notes can enable new research and support interoperability. For example, ‘severe fluid retention (edema)’ is not listed as an adverse effect of Tacrolimus, an anti-rejection drug, and only 135 out of more than 72,000 adverse events in the FDA’s FAERS repository [15] relate to edema. Although we do not claim a clinical association here, this type of scenario illustrates how important details may be difficult to identify when they are not consistently recorded in structured EHR fields. Highlighting the clinical notes of post-transplant patients who were prescribed Tacrolimus could facilitate exploratory analyses by making it easier to locate the various terms clinicians might use to describe edema (e.g., swollen feet) and to support hypothesis-generation for future studies.

To achieve effective highlighting of EHRs that meets the stated objectives, a comprehensive interface terminology is required. Although SNOMED CT [16] is the most comprehensive reference clinical terminology available, it does not capture the full richness and variability of the language used in cardiology EHRs. Many clinically important expressions, procedural nuances, and fine-granularity concepts are absent from structured terminologies. As a result, highlighting EHRs with SNOMED CT alone leaves large portions of clinically relevant detail unmarked, limiting its usefulness for fast skimming, clinical review, downstream applications such as summarization and simplification. A dedicated interface terminology is therefore needed to bridge the gap between the structured vocabulary of SNOMED CT and the granular, free-text descriptions that clinicians routinely write.

To address this limitation, we draw on the concept of an interface terminology [17-20]. Unlike a reference terminology, which provides a standardized, hierarchical vocabulary intended for coding, billing, and interoperability, an interface terminology is curated for a specific application (e.g., for highlighting EHRs) and focuses on reflecting the actual language clinicians use in practice.

In order to address the established need, we propose the design of a Cardiology Interface Terminology (CIT) dedicated to accurately highlighting EHRs of cardiology patients. The CIT interface terminology includes fine-granularity phrases, abbreviations, and multi-word concepts that do not appear in reference terminologies but are essential for capturing clinical meaning as written. For example, while SNOMED CT may contain the concept "sigmoidoscopy," CIT may additionally include expressions such as "flexible sigmoidoscopy" or "urgent repeat sigmoidoscopy," enabling more precise highlighting of clinical details. In the first phase, to create the initial CIT, we begin by deriving concepts from cardiology-related hierarchies in SNOMED CT. The initial CIT is further enriched by adding several important components, such as medical abbreviations, medications, and other necessary elements, since early research without such components did not yield satisfactory highlighting[21, 22]. Next, in the second phase, using an iterative, semi-automatic approach, we extract fine granularity phrases containing initial CIT concepts from EHRs, which are not already included in initial CIT, as CIT candidate concepts. All candidate concepts undergo both automatic and manual review to determine whether they fit the terminology before being added to the CIT. This process is repeated until a convergence criterion is met.

In the third phase of CIT curation, an innovative machine learning (ML) technique is employed to extract additional concepts from EHRs, further enriching the CIT. The output from the second phase serves as training data for the ML model used in the third phase. All of the concepts in CIT serve as positive samples in the training dataset, while rejected phrases serve as negative samples. We train a neural network (NN) model that classifies newly generated phrases to identify whether a phrase could be a CIT concept, and phrases labeled "1" are added to CIT. Finally, the resulting CIT from phase three is used to highlight a test set, evaluating how effectively the final CIT captures details in an unseen dataset during the curation of CIT.

# 2 Methods

First, we present the evaluation metrics and the data used in our analysis before describing the highlighting process.

## 2.1 Evaluation metrics

To evaluate how effectively the terminology captures clinically meaningful information, we use four metrics: coverage, breadth, completeness, and conciseness. In this context, a detail is defined as any clinically meaningful unit, such as a diagnosis, symptom, lab value, medication, procedure, or modifier, that is relevant for interpreting the clinical note.

Let

- $N_t$: total number of words in the note
- $N_h$: number of highlighted words
- $N_{hc}$: number of highlighted concepts
- $N_{td}$: total number of clinically relevant items of details
- $N_{hd}$: number of items of details correctly highlighted by the terminology
- $N_{th}$: total number of items of details highlighted by the terminology (including over-highlighted or irrelevant ones)

Coverage refers to the percentage of all highlighted words in a note, and is calculated using equation (1).

$$Coverage = \frac{N_h}{N_t} \times 100 \qquad (1)$$

Higher coverage indicates that a larger portion of a note's content is captured through the highlighting.

Breadth refers to the average number of words per highlighted concept, and is calculated by equation (2).

$$Breadth = \frac{N_h}{N_{hc}} \qquad (2)$$

A higher breadth implies that the concepts are longer and more granular, thus more accurately capturing the clinical information.

Completeness refers to the proportion of the details that are highlighted out of all details and is calculated by (3).

$$Completeness = \frac{N_{hd}}{N_{td}} \times 100 \qquad (3)$$

Conciseness refers to the proportion of highlighted details that should be highlighted (i.e., do not represent noise or over-highlighting) and is calculated by (4).

$$Conciseness = \frac{N_{hd}}{N_{th}} \times 100 \qquad (4)$$

## 2.2 Data

MIMIC-III is a publicly available database containing deidentified EHRs from over 40,000 patients, including demographics, clinical data, caregiver notes, imaging reports, and post-discharge information [23].

In our study we used MIMIC-III discharge notes of patients from two cardiology related ICUs: the Coronary Care Unit (CCU) and the Cardiac Surgery Recovery Unit (CSRU). We utilized two separate datasets, each containing 500 randomly selected MIMIC-III cardiology patient discharge notes. One, the build set B, is used for constructing the CIT, while the other, test set T, is employed to evaluate how well the highlights by CIT generalize to other notes. Due to their summarizing nature, discharge notes reflect all parts of EHRs.

## 2.3 Process

Our method involves a three-phase process. In Phase 1, we build the initial CIT by extracting concepts from cardiology-related hierarchies of SNOMED CT and enriching it with additional components such as abbreviations, medications, numbers, verbs, and other necessary elements. At the end of Phase 1, this initial CIT serves as the starting terminology for the next phase. Beginning with Phase 2, we refer to this terminology simply as CIT. In Phase 2, we iteratively mine fine-granularity phrases from EHRs and review them before adding them to CIT. In Phase 3, we further enrich CIT using a machine learning model trained on the accepted and rejected phrases from Phase 2.

Throughout this work, we use the term “concept” in a broad sense to reflect the different types of elements incorporated into CIT. To avoid ambiguity, we distinguish three categories: (1) standardized ontology-based concepts taken from SNOMED CT, which represent clinically defined entities; (2) mined concepts, which are multi-word expressions extracted from EHRs because they convey clinically meaningful information that does not appear in reference terminologies; and (3) auxiliary components such as verbs, abbreviations, negations, numbers, and measurement units, which support accurate highlighting but are not clinical entities themselves.

### 2.3.1 Phase 1: Building Initial Cardiology Interface Terminology

For creating the Initial CIT, we begin by adding the standardized ontology-based concepts that we extracted from 11 cardiology-related sub-hierarchies of SNOMED CT. Next, we add the standardized ontology-based concepts that appear in build set but not in initial CIT; these are not cardiology-related, but appear often in the EHRs of cardiology patients. To do that, first, we highlight build set with SNOMED CT, then we highlight build set with initial CIT. The standardized ontology-based concepts that are highlighted by SNOMED CT but are not highlighted with initial CIT, are identified as DIFF concepts.

All stop words [24] such as "as," "by," "with," etc., are removed from the DIFF concepts because such words act as connectors and should not serve as base concept for expansion. Remaining DIFF concepts are added to initial CIT.

Reviewing the EHRs in build set highlighted with this initial version of initial CIT, we observe that some necessary information such as abbreviations, medications, verbs, negation terms, numbers, which are not standardized ontology-based concepts in SNOMED CT, are not highlighted. Therefore, we add such concepts to initial CIT. Early research without such components did not yield satisfactory highlighting [21]. We explore each type of such auxiliary components in the following.

**Abbreviations:** Lists of abbreviations in cardiology [25], and in medicine in general [26, 27], and also those found in EHRs, have been inserted into initial CIT. Moreover, all abbreviations from build set, which did not appear in initial CIT or in these lists, have been extracted and added to initial CIT.

**Numbers:** We analyzed all the numbers in 500 EHRs of build set to determine what kind of numbers are expected in EHRs of cardiology patients. We encounter integers as age or test results e.g., blood pressure test, temperature readings and fractions as dosages. As a result, we have added the following types of numbers to initial CIT:

- All integers in 0-250, the integers 300,400,500,600, 700, 800, 900, 1000, 2000, 3000, 4000, 5000, 6000, 7000, 8000, 9000, 10000
- Decimal numbers between 0 and 102.9, with one decimal place.
- Fractions like 0.125, 0.15, 0.25, 0.375, 0.625, and 0.75.

- Emergency numbers 811 and 911.
- Words for "hundred," "thousand," "million," and "billion" in both singular and plural forms.
- Numbers written as words from "zero" to "nineteen," and the tens up to "ninety." ("Zero, One, …, nineteen, twenty, thirty…, ninety")
- Greek numerals for the numbers 1 to 10 (I, II, …).

**Negations:** Highlighting negations is crucial as they change the meaning of a statement; for example, highlighting only "diarrhea" in "No diarrhea" can have serious ramifications for clinical decisions, so "No" must also be highlighted for clarity. By analyzing EHRs in build set, we have created a list consisting of all the various forms of negations [28] that should be added to initial CIT.

**Medications:** All the medications commonly used in Cardiology [29] have been added to ICIT, both brand and generic names.

**Measurements:** All mass measurement units such as ml, cc, gr, gram, and liter have been added to initial CIT.

**Verbs**: Highlighting verbs in EHRs is important because verbs primarily convey actions the understanding of which is crucial. We have added 1000 common English verbs [30] and their variations to initial CIT. These variations include the base form, past form, past perfect form, third person form, and continuous form, have been added as synonyms with the same ID.

**Adding singular nouns to plurals and vice versa:** The goal is to ensure that both forms (pleural and singular) are highlighted. Identifying whether a concept is singular or plural requires knowing its part of speech in a sentence. However, this was not possible for the initial SNOMED CT concepts extracted from 11 cardiology-related hierarchies of SNOMED CT for lack of context, but it was possible for DIFF concepts. Using the NLTK library [31] , single-word DIFF concepts were identified as singular ('NN') or plural ('NNS'). The Inflect library was then used to generate the corresponding plural or singular forms, which were added to initial CIT as synonyms under the same ID. Each generated form was manually reviewed for accuracy before inclusion.

**Adding adjectives for adverbs and vice versa:** The methodology for mapping adjectives to their corresponding adverbs and vice versa in the initial CIT is as follows: Using DIFF concepts from build set with available original sentences, we identified the part of speech for each concept. Due to the lack of direct conversion tools or library and the limitations of WordNet in NLTK [31], we collaborated with a linguist (LL) to develop specific rules to establish these relationships, as standard linguistic patterns do not universally apply.

We developed a list of irregular adverbs for direct conversions. Regular adjectives ending in "y" are converted to adverbs by substituting "y" with "-ily," and for other adjectives, we append "ly." To derive adjectives from adverbs, we replace "-ily" with "y" or remove "ly." Adverbs not ending in "ly" and not listed as irregular typically lack a corresponding adjective. All entries were rigorously reviewed for accuracy before inclusion in initial CIT.

Fig. 1(b) shows an EHR note highlighted by initial CIT with coverage of 54.51% and breadth of 1.13.

#### 2.3.2 Phase 2: Enriching CIT with fine-granularity concepts mined from EHRs

We aim to create a comprehensive interface terminology by identifying and incorporating in it high-granularity phrases. The high-granularity phrases capture essential content, such as complex medical terms, many of which may not be included in existing medical terminologies like SNOMED CT. As shown in Fig. 1(b), while initial CIT contains many detailed concepts, it does not always cover all the granular phrases that clinicians use in EHRs, which are often necessary for accurately capturing patient information.

To address this gap, our goal is to expand the CIT by mining these missing phrases directly from the EHRs. Manually identifying such phrases requires domain experts and is time-consuming, expensive and inefficient, so we employ automated operations to extract them. These operations, concatenation and anchoring, are designed to systematically mine high-granularity phrases from EHRs by leveraging existing CIT concepts contained in such phrases.

Before applying concatenation or anchoring, we highlight the build set with initial CIT. Hence, we highlight standardized ontology-based concepts of SNOMED CT or axillary components.

**Concatenation:** Concatenation involves combining two or more adjacent highlighted concepts from the EHRs to form a new phrase. If stop words [24] are present between the adjacent concepts, they are included in the concatenated phrase. All the phrases generated by concatenation are automatically and manually reviewed for their fit as a valid mined concept of CIT.

Concatenation is applied based on Rule 1:

***Rule* 1:** $[existing_concept1]$ + $sw*$ + $[existing_concept2]$

where $[existing_concept1]$ and $[existing_concept2]$ are two concepts in the current terminology, and $sw*$ denotes zero or more stop words. (* indicates Kleene star [32])

For example, in Fig. 1(b), the two concepts "high" and "TSH" can be concatenated to generate the phrase "high TSH." After both automatic and manual review, this phrase will be added to the CIT as a legitimate mined concept. Another example involves concatenating the concepts "several" and "large," resulting in the phrase "several large." However, this phrase will be rejected automatically; it ends with an adjective, indicating that the phrase is incomplete.

**Anchoring:** Anchoring identifies an unhighlighted word adjacent to a highlighted concept, in the EHR note, on the left, right, or both sides of the concept to form a new phrase. Similar to concatenation, stop words between the highlighted concept and the unhighlighted word are allowed. As for concatenation, the resulting anchored phrases are also subjected to automatic and manual review to evaluate their fit as a valid mined concept of CIT.

Anchoring is applied based on Rule 2-4:

***Rule* 2:** $W1$ + $sw*$ + $[existing_concept]$

***Rule* 3:** $[existing_concept]$ + $sw*$ + $W1$

***Rule* 4:** $W1$ + $sw*$ + $[existing_concept]$ + $sw*$ + $W2$

where $[existing_concept]$ is a concept in the current terminology, $W1$ and $W2$ are unhighlighted words, and $sw*$ denotes zero or more stop words.

For example, in Fig. 1(b), the standardized ontology-based concept "sigmoidoscopy" can be anchored to its adjacent word on the left, "flexible," to generate the phrase "flexible sigmoidoscopy," which is a legitimate concept. In another example, the standardized ontology-based concept "thyroid" can be anchored to the adjacent word on its left, "based," and by including the stop word "on," we obtain the candidate mined concept "based on thyroid." This candidate mined concept will be rejected as a truncated phrase, since the context reveals that the full phrase is "based on thyroid studies". Additional examples of legitimate mined concepts and illegitimate phrases generated through concatenation and anchoring are shown in Table 1.

**Iterations**: We begin the process of mining fine-granularity concepts by highlighting build set with initial CIT. Next, we automatically extract all phrases obtained by concatenation of two or more concepts from initial CIT within the highlighted build set. These extracted phrases are then automatically and manually reviewed for their suitability as CIT mined concepts. In the automatic review, the structural validity of each mined phrase is checked using predefined rules and the part-of-speech (PoS) tag of the final word. For example, if a phrase ends with an adjective, adverb, or preposition, it is considered truncated and structurally invalid. A phrase such as "remained hemodynamically" ends with an adverb and is therefore rejected automatically. All automatically rejected phrases are added to a file referred to as the reject list. Phrases that pass the automatic review undergo manual review performed by two PhD students in computer science with extensive experience in medical informatics research. During manual review, each phrase is evaluated semantically in the context of the sentence from which it was extracted. For example, the phrase "several large BRBPR" is rejected because the correct concept in the context is "several large BRBPR transfusions." Accepted phrases are added to CIT, forming a new version (V1.1) of CIT, and manually rejected phrases are also appended to the reject list.

In subsequent iterations, newly mined phrases from concatenation or anchoring are first checked against the reject list to avoid re-reviewing phrases previously deemed invalid, and against the CIT to prevent redundant review of phrases that have already been reviewed and accepted. During the initial batches of manual review, both reviewers independently labeled the same phrases, compared their decisions, and resolved disagreements together. Once consistent labeling criteria were established, the reviewers divided the phrase lists and each reviewed half of the phrases. Inter-rater consistency was not calculated.

Subsequently, we highlight build set using the latest version ($V_{1.1}$) of CIT and automatically mine all phrases obtained by anchoring operation from the highlighted build set with the latest version of CIT. These phrases are also automatically and manually reviewed, and approved mined concepts are incorporated into CIT to generate a new version ($V_{1.2}$) of CIT, and rejected phrases are added to the reject list.

This process, comprising one round of concatenation followed by one round of anchoring, constitutes a single iteration. We repeat this iterative process until convergence is achieved. The criterion for convergence is that if the increased coverage of the last iteration is less than 2%, it indicates that further increase in coverage in future iterations will be insignificant, therefore we stop the iterations (see more about convergence in Discussion). Otherwise, we perform one more iteration by concatenation and anchoring. Fig. 1(c) shows the same note highlighted by CIT obtained in phase 2 after two iterations with coverage of 62.1% and breadth of 1.83.

### 2.3.3 Phase 3: Enriching CIT with Machine Learning techniques

Phase 3 focuses on enhancing CIT through the application of ML techniques. Phase 2 produced the training data required for Phase 3, consisting of: (a) 63,533 CIT concepts as the positive samples, and (b) the reject list (R), comprising 46,138 phrases that were rejected either automatically or manually, as negative samples. The training dataset phrases are embedded [33] using Clinical BioBERT [34], since it has been pretrained on a large corpus of clinical text, including approximately two million notes of the MIMIC-III database, making it a good choice for this study [35].

The training dataset is randomly divided into two parts: 80% allocated for training and validation, and 20% reserved for testing. A Neural Network (NN) [36] model was selected for training due to its demonstrated superiority in classification tasks compared to other classifiers [37]. After conducting a grid search [38] to fine-tune the hyperparameters, we ended up having a model configuration with a single hidden layer containing 100 neurons, using the Relu activation function [39] and the Adam optimizer [40]. To further enhance the model's performance, we employed a 5-fold cross-validation [41] during training. Additionally, a dropout rate [42] of 0.2 was applied to prevent overfitting. The resulting NN achieved an accuracy of 86%, along with precision, recall, and F1 scores of 88% each, when evaluated on the test set.

To enhance CIT, we extracted all subsequences of up to L words from each sentence in the 500 EHRs of build set, where L represents the maximum number of words allowed in the extracted phrases. If a sentence contained fewer than L words, the subsequences included only up to X words, where X represents the number of words in the sentence. Thus, the total number of phrases generated was proportional to L and the total number of words in build set, making the complexity of phrase generation linear with the dataset's word count. In this study, we tested L value of 6.

The new generated phrases, which are not already present as either positive (CIT) or negative (R) samples, were extracted from B. Additional rules, using NLTK library [43], were applied to eliminate phrases ending with adjectives, adverbs, or stop words. The remaining phrases were embedded using Clinical BioBERT. Then, we used our previously trained Neural Network model to classify the embedded phrases. Those classified as "1" (valid concepts) were added to CIT, resulting in the enhanced terminology, $CIT_{ML}$. Fig. 1(d) shows the same note highlighted with $CIT_{ML}$ coverage of 68.01% and breadth of 1.8.

For the evaluation of the EHRs in test set (T), $CIT_{ML}$ was further enriched with concepts from SNOMED CT that appeared in test set but were absent from $CIT_{ML}$, which we refer to as DIFF'. DIFF' concepts are added to $CIT_{ML}$, yielding $CIT_{ML+}$. Fig. 2 illustrates construction of CIT.

## 2.4 Extracting and highlighting headers

Headers are typically used to categorize text and further enhance readability. Reviewing EHRs in build set, we observe that most of the time, phrases followed by a ':' are considered headers. Following this pattern, we extracted headers from build set and counted how many times each was repeated. The resulting list, which includes headers and their frequency in build set, was reviewed by an internist (G.E.). Based on the outcome of this review, those repeated twice or more were classified as headers recommended for emphasizing. As shown in Fig. 1 (d), these headers are highlighted in pink and start a new line. This approach supports orientation of EHRs.

Fig. 1 shows various highlighting of a note in build dataset in (a) highlighted by SNOMED CT, with coverage of 31.41% and breadth of 1.24., in (b) highlighting by initial CIT with coverage of 54.51% and breadth of 1.13, in (c) highlighting by CITV2.2 with coverage of 62.1% and breadth of 1.83, and in (d) highlighting by $CIT_{ML}$ with coverage of 68.01% and breadth of 1.8. Red circles represent examples of mined concepts obtained through concatenation, while green circles represent examples of mined concepts obtained through anchoring. Black underlines indicate examples of mined concepts derived iteratively using both concatenation and anchoring. Yellow circles in (d) represent examples of mined concepts mined exclusively via the ML model.

# 3 Results

Table 1 shows examples of legitimate mined concepts and illegitimate phrases form build set generated through concatenation and anchoring.

Legitimate mined concepts generated through concatenation are used in the anchoring process to create finer granularity phrases. Similarly, mined concepts produced via anchoring are utilized in the subsequent concatenation iteration to generate more refined phrases. For example, in Fig. 1(b), the concepts "RBC" and "transfusions" are first concatenated to generate "RBC transfusions." Since this is a valid mined concept, it is added to the CIT. Next, in the anchoring process, "RBC transfusions" is anchored to the adjacent word on its left, "further," resulting in the phrase "further RBC transfusions." As this is also a legitimate concept, it is added to the CIT. In the following concatenation iteration, "further RBC transfusions" is concatenated with the concept "minimal," which generates the fine-granularity mined concept "minimal further RBC transfusions." This iterative process ensures that the terminology evolves to capture relevant and meaningful concepts from EHRs efficiently. Additional examples of fine-granularity mined concepts from build set generated through the iterative process of concatenation and anchoring are shown in Table 2.

He had several episodes of hematochezia at his nursing home today, and was subsequently sent in to the ED. He denied any chest pain, SOB, lighthededness, diarrhea, abdominal pain, fevers, chills, back pain. . Last week he also presented with BRBPR, and he was discharged yesterday. Initially, a tagged red blood cell scan showed bleeding from the rectum. However, subsequent angiography was unable to localize and embolize the source of bleed. A flexible sigmoidoscopy showed an actively bleeding Dielafoys lesion which was successfully clipped. The patient subsequently remained hemodynamically stable, with minimal further RBC transfusions. Coumadin, aspirin, and Plavix were initially held. Aspirin and Plavix were resumed prior to discharge. ED course: He was found to have a stable hct from last admission. He had several large BRBPR episodes in ED, and his SBP went from 140 to 110. He was give 1 liter of NS. GI was called, and planned for scope in AM. He was admitted to MICU for close observation 1.Severe PVD: His wound vac was replaced while in the hospital. AFib: The patient carries a history of atrial fibrillation, status post ablation. He has been stable in sinus rhythm throughout his hospitalization. His coumadin was stopped during his previous hospitalization. We continued to hold his coumadin in light of his GI bleeding. Hypothyroidism: His outpatient thyroid replacement was continued. Based on thyroid studies from previous admission (high TSH, low T4), the patient may require titration of his medications when not in an acute setting. HTN: In the setting of acute bleeding, antihypertensives were initially held, prior to discharge the patients home dose of Imdur and metoprolol.

(a)

He had several episodes of hematochezia at his nursing home today, and was subsequently sent in to the ED. He denied any chest pain, SOB, lighthededness, diarrhea, abdominal pain, fevers, chills, back pain. . Last week he also presented with BRBPR, and he was discharged yesterday. Initially, a tagged red blood cell scan showed bleeding from the rectum. However, subsequent angiography was unable to localize and embolize the source of bleed. A flexible sigmoidoscopy showed an actively bleeding Dielafoys lesion which was successfully clipped. The patient subsequently remained hemodynamically stable, with minimal further RBC transfusions. Coumadin, aspirin, and Plavix were initially held. Aspirin and Plavix were resumed prior to discharge. ED course: He was found to have a stable hct from last admission. He had several large BRBPR episodes in ED, and his SBP went from 140 to 110. He was give 1 liter of NS. GI was called, and planned for scope in AM. He was admitted to MICU for close observation 1.Severe PVD: His wound vac was replaced while in the hospital. AFib: The patient carries a history of atrial fibrillation, status post ablation. He has been stable in sinus rhythm throughout his hospitalization. His coumadin was stopped during his previous hospitalization. We continued to hold his coumadin in light of his GI bleeding. Hypothyroidism: His outpatient thyroid replacement was continued. Based on thyroid studies from previous admission (high TSH, low T4), the patient may require titration of his medications when not in an acute setting. HTN: In the setting of acute bleeding, antihypertensives were initially held, prior to discharge the patients home dose of Imdur and metoprolol.

(b)

He had several episodes of hematochezia at his nursing home today, and was subsequently sent in to the ED. He denied any chest pain, SOB, lighthededness, diarrhea, abdominal pain, fevers, chills, back pain. . Last week he also presented with BRBPR, and he was discharged yesterday. Initially, a tagged red blood cell scan showed bleeding from the rectum. However, subsequent angiography was unable to localize and embolize the source of bleed. A flexible sigmoidoscopy showed an actively bleeding Dielafoys lesion which was successfully clipped. The patient subsequently remained hemodynamically stable, with minimal further RBC transfusions. Coumadin, aspirin, and Plavix were initially held. Aspirin and Plavix were resumed prior to discharge. ED course: He was found to have a stable hct from last admission. He had several large BRBPR episodes in ED, and his SBP went from 140 to 110. He was give 1 liter of NS. GI was called, and planned for scope in AM. He was admitted to MICU for close observation 1.Severe PVD: His wound vac was replaced while in the hospital. AFib: The patient carries a history of atrial fibrillation, status post ablation. He has been stable in sinus rhythm throughout his hospitalization. His coumadin was stopped during his previous hospitalization. We continued to hold his coumadin in light of his GI bleeding. Hypothyroidism: His outpatient thyroid replacement was continued. Based on thyroid studies from previous admission (high TSH, low T4), the patient may require titration of his medications when not in an acute setting. HTN: In the setting of acute bleeding, antihypertensives were initially held, prior to discharge the patients home dose of Imdur and metoprolol.

(c)

He had several episodes of hematochezia at his nursing home today, and was subsequently sent in to the ED. He denied any chest pain, SOB, lighthededness, diarrhea, abdominal pain, fevers, chills, back pain. . Last week he also presented with BRBPR, and he was discharged yesterday. Initially, a tagged red blood cell scan showed bleeding from the rectum. However, subsequent angiography was unable to localize and embolize the source of bleed. A flexible sigmoidoscopy showed an actively bleeding Dielafoys lesion which was successfully clipped. The patient subsequently remained hemodynamically stable, with minimal further RBC transfusions. Coumadin, aspirin, and Plavix were initially held. Aspirin and Plavix were resumed prior to discharge.

ED Course: He was found to have a stable hct from last admission. He had several large BRBPR episodes in ED, and his SBP went from 140 to 110. He was give 1 liter of NS. GI was called, and planned for scope in AM. He was admitted to MICU for close observation 1.Severe PVD: His wound vac was replaced while in the hospital. AFib: The patient carries a history of atrial fibrillation, status post ablation. He has been stable in sinus rhythm throughout his hospitalization. His coumadin was stopped during his previous hospitalization. We continued to hold his coumadin in light of his GI bleeding. Hypothyroidism: His outpatient thyroid replacement was continued. Based on thyroid studies from previous admission (high TSH, low T4), the patient may require titration of his medications when not in an acute setting. HTN: In the setting of acute bleeding, antihypertensives were initially held, prior to discharge the patients home dose of Imdur and metoprolol.

(d)

**Figure 1.** Highlights of a note in build set by (a) SNOMED CT (b) initial CIT (c) $CIT_{V2.2}$ (d) $CIT_{ML}$

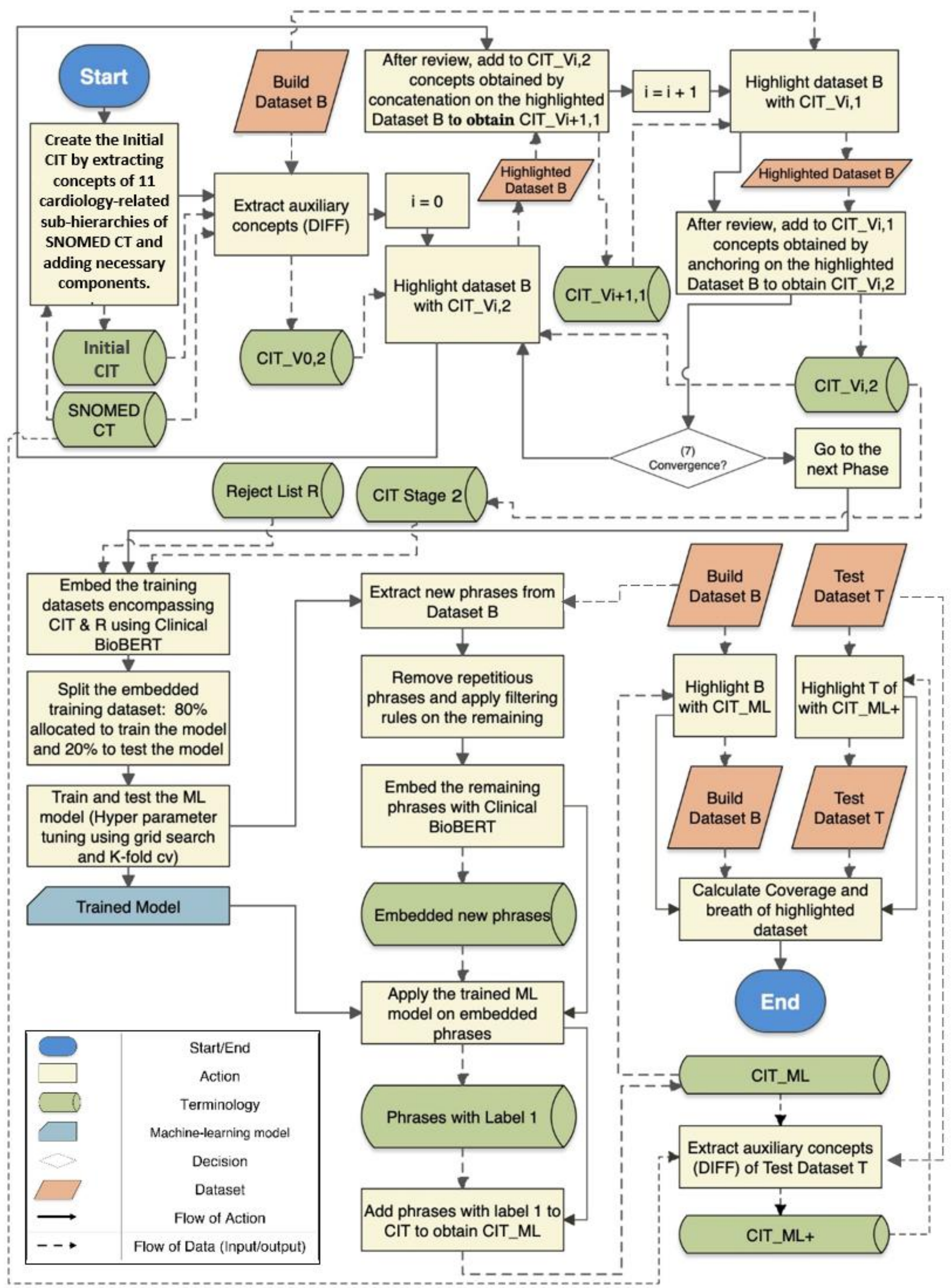

**Figure 2.** CIT Construction

**Table 1.** Legitimate mined concepts and illegitimate phrases containing initial CIT concepts, shown in bold, generated by concatenation or anchoring

| | **legitimate** | **illegitimate** |
|---|---|---|
| **Concatenation** | **status post** \| **ablation**<br>**high** \| **TSH**<br>**denied** \|any \|**chest pain**<br>**setting** \|of \|**acute bleeding** | **several** \| **large**<br>**give** \| **1**<br>**coumadin** \| in \| **light**<br>**studies** \| from \| **previous** |
| **Anchoring** | flexible \| **sigmoidoscopy**<br>**successfully** \| clipped<br>actively \| **bleeding**<br>Dielafoys \| **lesion** | Based \| on \| **thyroid**<br>**patient** \| subsequently<br>embolize \| the \| **source**<br>**sinus rhythm** \| throughout |

Table 3 illustrates the evolution of coverage and breadth metrics during the iterative construction of the CIT. Each iteration consists of one concatenation (denoted by the index ".1") followed by an anchoring operation (denoted by the index ".2"). For example, in the first iteration, in $V_{1.1}$ the first "1" represents the first iteration and the second "1" represents concatenation. Similarly, for $V_{1.2}$, “1” indicates the first iteration, and "2" represents anchoring.

At the end of the first iteration, the coverage reaches 69.08%, increasing to 71.00% after the second iteration. The improvement of 1.92% is below the 2% threshold set as our convergence criterion, indicating that convergence is achieved. Therefore, the process is stopped after the second iteration.

After convergence in phase 2, $CIT_{V2.2}$ was then further enriched to $CIT_{ML}$, incorporating the additional mined concepts extracted from build set in phase 3. On the other hand, for test data test set, we added the DIFF concepts of test set which are in SNOMED CT but not in $CIT_{ML}$, forming $CIT_{ML+}$. Highlighted test set accordingly with it. $CIT_{ML}$ achieved a coverage of 75.17% with a breadth of 2.05 for build set, while $CIT_{ML+}$ obtained a coverage of 74.22% and a breadth of 1.68 for T.

All in all, the average coverage with $CIT_{ML}$ for build set increased by 40.67% versus coverage with SNOMED CT (a relative increase of 117%), while with $CIT_{ML+}$ for test set, it increased by 38.92% (a relative increase of 110%) compared to SNOMED CT. The average breadth also increased by 0.81 versus breadth with SNOMED CT (a relative increase of 65.3%) for build set and by 0.43 (a relative increase of 34.4%) for T

These results for test set, which was not used during the curation of the CIT, demonstrate the efficacy, comprehensiveness and richness of the resulting CIT.

**Table 2.** The process of obtaining examples of fine-granularity mined concepts.

| | | | |
|---|---|---|---|
| | Initial Phrase | actively **bleeding** Dielafoys **lesion** | Based on **thyroid** **studies** from **previous** **admission** |
| 1 | Concatenation | -- | **thyroid** \| **studies**<br>**previous** \| **admission** |
| | Anchoring | Actively \| **bleeding**<br>Dielafoys \| **lesion** | Based \| on \| **thyroid studies** |
| 2 | Concatenation | **actively bleeding** \| **Dielafoys lesion** | **Based on thyroid studies** \| from \| **previous admission** |
| | Anchoring | -- | -- |
| Final mined concept | | **actively bleeding Dielafoys lesion** | **Based on thyroid studies from previous admission** |

**Table 3.** Coverage and Breadth for Build/Test sets

| **Iteration** | **Terminology** | **Terminology Size** | **#New Concepts added to CIT** | **#New Concepts Seen in Test Data T** | **Build Data B** | | **Test Data T** | |
|---|---|---|---|---|---|---|---|---|
| | | | | | **Coverage** | **Breadth** | **Coverage** | **Breadth** |
| – | SNOMED CT | 899,773 | – | – | 34.5% | 1.24% | 35.3% | 1.25 |
| – | initial CIT | 42,931 | – | – | 63.54% | 1.15 | – | – |
| 1 | $CIT_{V1.1}$ | 56,112 | 13181 | 5817 concepts appeared 12787 times | 64.50% | 1.66 | – | – |
| | $CIT_{V1.2}$ | 60,152 | 4040 | 1638 concepts appeared 2905 times | 69.08% | 1.78 | – | – |
| 2 | $CIT_{V2.1}$ | 63,193 | 3046 | 1432 concepts appeared 1927 times | 70.66% | 1.99 | – | – |
| | $CIT_{V2.2}$ | 63,533 | 340 | 122 concepts appeared 156 times | 71.00% | 2.00 | – | – |
| – | $CIT_{ML}$ | 75,132 | – | – | **75.17%** | **2.05** | – | – |
| | $CIT_{ML+}$ | 76,943 | – | – | – | – | **74.22%** | **1.68** |

As an example, Fig 3 shows different highlights by various terminologies for a note in test set, in (a) highlighted with SNOMED CT with coverage of 22.29% and breadth of 1.08, in (b) highlighted with initial CIT with coverage of 57.09% and breadth 1.05, in (c) highlighted with $CIT_V_{2.2}$ with coverage of 63.73% and breadth of 1.59, and in (d) highlighted with $CIT_{ML+}$ with coverage 71.18%, breadth 1.71, completeness of 97.2%, and conciseness of 84.9%. Red circles are the unhighlighted phrases that are highlighted with next versions of CIT in the next parts of the figure. Green circles are the highlighted mined concepts that should not be highlighted. The figure demonstrates a clear, gradual improvement in both coverage and breadth metrics as the CIT evolves, demonstrated by circling in red the unhighlighted terms that were highlighted with the next version of CIT, and by circling in green the highlighted mined concepts that should not be highlighted.

To evaluate the effectiveness of highlighting in the test set, we manually inspected 20 random notes and calculated the completeness and conciseness metrics. Table 4 shows the manually calculated completeness and conciseness for 20 notes in the test data, highlighted with $CIT_{ML+}$, along with their coverage.

# 4 Discussion

**Results:** We observe that the breadth for the highlights by initial CIT is smaller than that by SNOMED CT. This disparity is due to the fact that the DIFF set of concepts in initial CIT, derived from the DIFF operation between SNOMED CT and initial CIT highlights, consists of 'shorter' SNOMED CT concepts. Additionally, many of the components incorporated within initial CIT, such as verbs, adjectives, adverbs, and medications, are typically single-word entities.

Concatenation contributed modest increases in coverage because this operation combines currently highlighted concepts, often adding few stop words. However, concatenation significantly improved breadth. For example, the coverage for $CIT_{1.1}$ is reflecting an increase of 0.96% (a relative increase of 1.5%) while breadth improved by 0.51 (a relative increase of 44.3%). The reason is that concatenation merges short concepts into longer, more fine-grained mined concepts.

Anchoring, on the other hand, enhanced both metrics, coverage and breadth. This is achieved by highlighting additional concepts through the integration of unhighlighted words to the left, right, or both sides of the highlighted concepts. For example, from $CIT_{V1.1}$ to $CIT_{V1.2,}$ the coverage increased by 4.58% (a relative increase of 7.1%), while the breadth increased by 0.12 (a relative increase of 7.2%).

Although the coverage of dataset build set highlighted with $CIT_{ML}$ and the coverage of test set highlighted with $CIT_{ML+}$ are nearly identical, the breadth of highlights in test set is notably smaller compared to highlights in B. The reason is that $CIT_{ML}$ is curated by mining concepts from build set, followed by the generation of longer phrases through concatenation and anchoring of these concepts. Conversely, dataset test set was entirely unseen during the process, and thus, concatenation and anchoring techniques were not employed on its concepts.

**Table 4.** Completeness, conciseness, and coverage for 20 random notes in the test set

| EHR note | completeness | conciseness | coverage |
|---|---|---|---|
| 1 | 98.4% | 84.1% | 78.3% |
| 2 | 100% | 81.8% | 55.4% |
| 3 | 92.3% | 84.6% | 66.6% |
| 4 | 100% | 87.5% | 66.4% |
| 5 | 95.3% | 72.1% | 85.16% |
| 6 | 100% | 85.1% | 57.9% |
| 7 | 100% | 89.3% | 71.1% |
| 8 | 100% | 84.1% | 72% |
| 9 | 100% | 77.2% | 77.7% |
| 10 | 98.2% | 86.6% | 70.5% |
| 11 | 97.8% | 90.7% | 67.7% |
| 12 | 100% | 84.3% | 71.6% |
| 13 | 100% | 89.1% | 60.5% |
| 14 | 97.5% | 89.2% | 78.8% |
| 15 | 100% | 79.1% | 66.4% |
| 16 | 95% | 83.4% | 61.73% |
| 17 | 100% | 80.3% | 73.1% |
| 18 | 97.5% | 85.1% | 69.3% |
| 19 | 97.4% | 86.3% | 74.3% |
| 20 | 97.2% | 84.9% | 71.1% |
| **Average** | **98.4%** | **84.2%** | **69.7%** |

During the iterations, a mined concept formed by concatenation must include concepts previously obtained through anchoring; otherwise, it would have been created in an earlier concatenation. Similarly, a mined concept obtained through anchoring must incorporate a mined concept derived from a prior concatenation or anchoring. Consequently, the number of mined concepts in each iteration decreases significantly, explaining the convergence criterion. The number of mined concepts in subsequent iterations is expected to drop sharply, with minimal impact on coverage for B. This effect is further reduced for test set due to the low frequency of longer mined concepts from the CIT in the test set, as shown in an earlier study [22].

**Data analyses:** We experimented with a maximum length of 6 for newly extracted phrases from build set to be labeled by the ML model. This decision was based on the observation that only 3% of the mined concepts obtained via concatenation and anchoring have a length greater than 6 words. Moreover, as the mined concepts extracted from build set become longer, the probability of their repetition in test set decreases, eliminating the need to mine longer phrases from build set. In a preliminary study [22], we experimented with both six-word and nine-word lengths. The difference for test set, the dataset that is indeed important for demonstrating the effectiveness of CIT, was only about 1% in coverage and 0.02 in breadth, indicating that conducting the experiment with nine words was unnecessary.

$CIT_{ML+}$ is envisioned as a dynamic terminology rather than a static one. This implies that $CIT_{ML}$ can be adaptability enriched to adjust to new sources of EHRs, such as those from a specific healthcare provider. By integrating all DIFF concepts from the new source (the SNOMED CT concepts that are in the EHRs of the new source but not in $CIT_{ML}$), we can enrich $CIT_{ML}$ to a $CIT_{ML+}$ that bridges the gap between the $CIT_{ML}$ and SNOMED CT for the new repository.

**Performance:** We conducted experiments on a NN model with both single and multiple hidden layers. Our findings indicate that increasing the depth of the neural networks did not enhance performance beyond that achieved by simpler NN structures. This may be attributed to the requirement for more extensive training data to improve the performance of deeper NNs. Future experiments will involve larger datasets to further investigate this hypothesis.

To enrich the CIT, we have tested different approaches. For example, we also constructed the CIT using the NER approach [44]; however, we obtained better results with the ML method.

**Evaluation metrics:** In our method, standardized ontology-based concepts and axillary components are carefully expanded and evaluated before being added to CIT. Extracted phrases are included in CIT only if they meet specific criteria for relevance, both semantically and structurally. So, the coverage metric indicates the amount of details of a note that should not be missed. Additionally, when concepts that are adjacent and closely related to each other are highlighted separately, readers must exert mental effort to connect them. To reduce this effort, following a cardiologist (A.E.) recommendation, we prioritize longer phrases that capture high-granularity concepts and preserve proper semantics. This is where the metric of breadth becomes important: greater breadth reduces the cognitive load required to skim the note, making it easier for readers to process the information efficiently.

60 yo M PMH sig for sz disorder presented with onset of severe, burning/heavy substernal CP and SOB. He had no N/V. Pain started at 1:00 pm and pt presented. Unclear if this was due to cath procedure itself or if the pt had ongoing ischemia. Wedge in cath lab was 24. Pt has age and smoking as traditional risk factors for MI - FH and cholesterol unknown. No known HTN Sz d/o s/p head injury at age 16 - last sz at age 18. BRIEF OVERVIEW: The patient is a 60 yo with a h/o sz d/o who presented from with CP and STEMI (IMI). Found to have focal lesions in 3 vessels. Stented x 2 in RCA with very little change in EKG. Cath also revealed LCx and LAD lesions. During the cath procedure, after stenting, the patient had a vagal episode with decreased BP and HR and was given atropine. HR and BP returned, but pt had 8/10 chest pain. Swan showed elevated BP, right sided pressures, and wedge. The pain subsided and the patient was given lasix and nitro in the lab and BB in the CCU (where he was taken for monitoring). He then became hypotensive and was symptomatic with nausea and LH. Swan showed decreased wedge and R sided pressures. He was placed in trendelenberg position and given fluids to bring up his BP. It was though that his low BP was due to decreased preload in a preload dependent state. Echo at the bedside revealed a hypokinetic free RV wall. After fluids and time for meds to wear off, the patient was normotensive and felt well. ASA and plavix were continued. Integrellin was d/cd after 18 hours per protocol. The remainder of his recovery was uneventful.

(a)

60 yo M PMH sig for sz disorder presented with onset of severe, burning/heavy substernal CP and SOB. He had no N/V. Pain started at 1:00 pm and pt presented. Unclear if this was due to cath procedure itself or if the pt had ongoing ischemia. Wedge in cath lab was 24. Pt has age and smoking as traditional risk factors for MI - FH and cholesterol unknown. No known HTN Sz d/o s/p head injury at age 16 - last sz at age 18. BRIEF OVERVIEW: The patient is a 60 yo with a h/o sz d/o who presented from with CP and STEMI (IMI). Found to have focal lesions in 3 vessels. Stented x 2 in RCA with very little change in EKG. Cath also revealed LCx and LAD lesions. During the cath procedure, after stenting, the patient had a vagal episode with decreased BP and HR and was given atropine. HR and BP returned, but pt had 8/10 chest pain. Swan showed elevated BP, right sided pressures, and wedge. The pain subsided and the patient was given lasix and nitro in the lab and BB in the CCU (where he was taken for monitoring). He then became hypotensive and was symptomatic with nausea and LH. Swan showed decreased wedge and R sided pressures. He was placed in trendelenberg position and given fluids to bring up his BP. It was though that his low BP was due to decreased preload in a preload dependent state. Echo at the bedside revealed a hypokinetic free RV wall. After fluids and time for meds to wear off, the patient was normotensive and felt well. ASA and plavix were continued. Integrellin was d/cd after 18 hours per protocol. The remainder of his recovery was uneventful.

(b)

60 yo M PMH sig for sz disorder presented with onset of severe, burning/heavy substernal CP and SOB. He had no N/V. Pain started at 1:00 pm and pt presented. Unclear if this was due to cath procedure itself or if the pt had ongoing ischemia. Wedge in cath lab was 24. Pt has age and smoking as traditional risk factors for MI - FH and cholesterol unknown. No known HTN Sz d/o s/p head injury at age 16 - last sz at age 18. BRIEF OVERVIEW: The patient is a 60 yo with a h/o sz d/o who presented from with CP and STEMI (IMI). Found to have focal lesions in 3 vessels. Stented x 2 in RCA with very little change in EKG. Cath also revealed LCx and LAD lesions. During the cath procedure, after stenting, the patient had a vagal episode with decreased BP and HR and was given atropine. HR and BP returned, but pt had 8/10 chest pain. Swan showed elevated BP, right sided pressures, and wedge. The pain subsided and the patient was given lasix and nitro in the lab and BB in the CCU (where he was taken for monitoring). He then became hypotensive and was symptomatic with nausea and LH. Swan showed decreased wedge and R sided pressures. He was placed in trendelenberg position and given fluids to bring up his BP. It was though that his low BP was due to decreased preload in a preload dependent state. Echo at the bedside revealed a hypokinetic free RV wall. After fluids and time for meds to wear off, the patient was normotensive and felt well. ASA and plavix were continued. Integrellin was d/cd after 18 hours per protocol. The remainder of his recovery was uneventful.

(c)

60 yo M PMH sig for sz disorder presented with onset of severe, burning/heavy substernal CP and SOB. He had no N/V. Pain started at 1:00 pm and pt presented. Unclear if this was due to cath procedure itself or if the pt had ongoing ischemia. Wedge in cath lab was 24. Pt has age and smoking as traditional risk factors for MI-FH and cholesterol unknown.No known HTN Sz d/o s/p head injury at age 16 - last sz at age 18. BRIEF OVERVIEW: The patient is a 60 yo with a h/o sz d/o who presented from with CP and STEMI (IMI). Found to have focal lesions in 3 vessels. Stented x 2 in RCA with very little change in EKG. Cath also revealed LCx and LAD lesions. During the cath procedure, after stenting, the patient had a vagal episode with decreased BP and HR and was given atropine. HR and BP returned, but pt had 8/10 chest pain. Swan showed elevated BP, right sided pressures, and wedge. The pain subsided and the patient was given lasix and nitro in the lab and BB in the CCU (where he was taken for monitoring). He then became hypotensive and was symptomatic with nausea and LH. Swan showed decreased wedge and R sided pressures. He was placed in trendelenberg position and given fluids to bring up his BP. It was though that his low BP was due to decreased preload in a preload dependent state. Echo at the bedside revealed a hypokinetic free RV wall. After fluids and time for meds to wear off, the patient was normotensive and felt well. ASA and plavix were continued. Integrellin was d/cd after 18 hours per protocol. The remainder of his recovery was uneventful.

(d)

**Figure 3.** Highlights of a note in test set by (a) SNOMED CT, (b) initial CIT, (c) CIT_V$_{2.2}$ (d) CIT$_{ML+}$

**Ideal highlighted text**: An EHR note is ideally highlighted when all and only the information that a medical professional would consider as important data is highlighted. Clinicians' writing styles and readers' needs vary, making it difficult to establish a single threshold to assess the desired amount of the highlighted text. Therefore, we need the opinions of MDs to assess how good a highlighting is. Although this assessment is also subjective, there is no better proxy to be used in this regard. Therefore, we requested G.E., an internist scientist experienced with EHRs, to manually highlight a random selection of 15 discharge notes from build set. We then evaluated the coverage of these manually highlighted records. The results revealed that the highest coverage was 82%, the lowest was 52%, and the average was 68%, with a standard deviation of 8.34. Consequently, we aim for an average coverage in the 68-78% range. Given that automatic highlighting tends to be less precise than manual highlight, we apply an upward standard deviation margin to minimize the risk of omitting crucial details, providing a reasonable approximation for effective highlighting.

**LLMs:** We tested highlighting the details in EHRs using ChatGPT-4o with prompt engineering. We utilized the 'Tone' strategy so that the model would act as a cardiologist. Following the best practices in [45, 46], we crafted and refined the prompt iteratively. Our final prompt is as follows:

*'Act as a cardiologist, and highlight the details in this EHR note. Keep the original text, and do not change it.'*

As an example, Fig 4 shows the EHR note of Fig 3, highlighted by ChatGPT-4o using the above prompt, where the bold words indicate the highlights.

However, many important pieces of information, including medical terms such as 'LCx and LAD lesions,' 'R sided pressures,' 'elevated BP,' and 'given lasix and nitro,' remain unhighlighted. In contrast, our automated approach, which uses interface terminology designed for highlighting EHRs, captures nearly all their details of this note.

Moreover, incorporating LLMs for highlighting EHRs poses ethical challenges [8, 47-49], including ensuring patient privacy through robust de-identification [50, 51], securing data against unauthorized access [51, 52], preserving data integrity to avoid misinterpretations [53-55], and addressing biases and reliability issues [56, 57], especially as LLMs may lack medical-specific training [58, 59].

**Limitations:** In this paper we proposed a semi-automated approach for developing an interface terminology specifically designed for highlighting of EHRs of cardiology patients. The rationale behind the limitation to cardiology is that using a single interface terminology to highlight general EHRs, similar to SNOMED CT reference terminology, would be too unwieldy to manage. While our focus is currently on cardiology, we aim to apply Transfer Learning [60-63] to create similar terminologies for other medical fields, such as Pulmonology.

**60 yo M PMH sig for sz disorder presented with onset of severe, burning/heavy substernal CP and SOB**. He had no N/V. **Pain started at 1:00 pm and pt presented**. Unclear if this was due to cath procedure itself or if the pt had ongoing ischemia. **Wedge in cath lab was 24**. **Pt has age and smoking as traditional risk factors for MI** - FH and cholesterol unknown. No known HTN **Sz d/o s/p head injury at age 16** - **last sz at age 18**, **BRIEF OVERVIEW: The patient is a 60 yo with a h/o sz d/o who presented from with CP and STEMI (IMI). Found to have focal lesions in 3 vessels. Stented x 2 in RCA with very little change in EKG**. Cath also revealed LCx and LAD lesions. During the cath procedure, after stenting, the patient had a vagal episode with decreased BP and HR and was given atropine. **HR and BP returned, but pt had 8/10 chest pain**. Swan showed elevated BP, right sided pressures, and wedge. The pain subsided and the patient was given lasix and nitro in the lab and BB in the CCU (where he was taken for monitoring). He then became hypotensive and was symptomatic with nausea and LH. Swan showed decreased wedge and R sided pressures. He was placed in trendelenberg position and given fluids to bring up his BP. It was thought that his low BP was due to decreased preload in a preload dependent state. Echo at the bedside revealed a hypokinetic free RV wall. After fluids and time for meds to wear off, the patient was normotensive and felt well. **ASA and plavix were continued. Integrellin was d/cd after 18 hours per protocol**. The remainder of his recovery was uneventful.

**Figure 4.** EHR note highlighted by LLMs

Manual review of phrases presents significant challenges, primarily due to its time-consuming and costly nature. However, an ML model can be trained to label most of the phrases (see Future Work). Moreover, manual reviews are inherently subjective, and the outcomes are influenced by the reviewers' personal judgments. To address this, we established rules for accepting or rejecting phrases in an interface terminology. For example, phrases ending with an adjective should be rejected. These rules can be updated and enriched further in the future to reduce the subjectivity of the reviewer's judgment.

Our method for mining fine-granularity concepts from EHRs relies on expanding existing highlighted concepts within the notes through concatenation and anchoring. The success of this process relies on having foundational concepts already included in the initial CIT. If these foundational concepts are missing, surrounding words cannot be anchored. A lack of comprehensive components in the initial CIT significantly limits our ability to extract fine-granularity concepts from EHRs. However, such concepts can be obtained using the ML model.

Additionally, over-highlighting in some EHRs can diminish the efficiency of our approach. This occurs when phrases of lesser importance, accepted by the judgment of the reviewer, are highlighted, thereby reducing the time-saving benefits of quickly skimming the highlighted notes. Although the average coverage of our test data is 74%, implying a 26% time savings, there are instances where greater efficiency might be achievable.

In Phase 2 of CIT curation, a large number of mined phrases must be manually reviewed to create the training dataset for the ML model used in Phase 3. This manual review process is time-consuming and costly. In our recent work [64], we proposed an ML-assisted approach to substantially reduce this effort. We trained a Neural Network (NN) model on varying subsets of phrases generated through concatenation and anchoring to determine the minimum amount of manual review needed to produce an effective training set. The results showed that manually reviewing only 25% of the concatenation-generated phrases and 30% of the anchoring-generated phrases was sufficient to train an ML model that could label the remaining phrases. When the resulting terminology was used to highlight the test set, the coverage, breadth, completeness, and conciseness metrics closely matched those obtained using the fully manually reviewed Phase 2 dataset. These findings demonstrate that the manual curation required for adapting this methodology to other medical domains can be reduced substantially without compromising highlighting quality, making the approach more scalable across diverse clinical specialties.

**Future works:** Future work will extend our approach to other specialties (e.g., pulmonology) through transfer learning, thereby reducing the extent of manual review. Additionally, we will explore downstream applications in which highlighting may improve model performance, including clinical question-answering systems.

# 5 Conclusion

In this paper, we introduced a semi-automated approach to develop a Cardiology Interface Terminology (CIT) tailored to highlight all details in EHRs of cardiology patients. Our innovative methodology leverages iterative processes of concatenation and anchoring, and a Neural Network model to systematically mine and refine concepts, addressing the limitations of existing reference terminologies such as SNOMED CT. The obtained CIT achieves significant improvements in completeness, conciseness, coverage and breadth, demonstrating its ability to capture almost all details in unseen datasets. By facilitating fast skimming, accurate AI summarization, and improved EHR simplification, this approach addresses both clinical health informatics and research challenges.

# 6 Funding

This research received no specific grant from any funding agency in the public, commercial, or not-for-profit sectors.

### References

1. Polepalli Ramesh, B., T. Houston, C. Brandt, H. Fang, and H. Yu, *Improving patients' electronic health record comprehension with NoteAid*, in *MEDINFO 2013*. 2013, IOS Press. p. 714-718.
2. Zhou S, Sen P, Liu H, Perl Y, Dehkordi MK. CFC annotator: a cluster-focused combination algorithm for annotating electronic health records by referencing interface terminology. In: Proceedings of the 18th International Joint Conference on Biomedical Engineering Systems and Technologies. 2025. Presented at: BIOSTEC 2025; February 19-21, 2025; Porto, Portugal. [doi: 10.5220/0013244500003911].
3. Zhou, S., M.K.H. Dehkordi, Y. Perl, F.P. Deek, and H. Liu, *Enhancing Electronic Health Records Annotation with a Cluster-Focused Combination Algorithm and Interface Terminologies.* Springer Book of HEALTHINF 2025.
4. Hassanzadeh, H., A. Nguyen, and B. Koopman. *Evaluation of medical concept annotation systems on clinical records*. in *Proceedings of the Australasian Language Technology Association Workshop 2016*. 2016.
5. Dymek, C., B. Kim, G.B. Melton, T.H. Payne, H. Singh, and C.-J. Hsiao, *Building the evidence-base to reduce electronic health record–related clinician burden.* Journal of the American Medical Informatics Association, 2021. **28**(5): p. 1057-1061.
6. Apathy, N.C., L. Rotenstein, D.W. Bates, and A.J. Holmgren, *Documentation dynamics: note composition, burden, and physician efficiency.* Health Services Research, 2023. **58**(3): p. 674-685.
7. Cui, S., J. Luo, M. Ye, J. Wang, T. Wang, and F. Ma. *MedSkim: Denoised Health Risk Prediction via Skimming Medical Claims Data*. in *2022 IEEE International Conference on Data Mining (ICDM)*. 2022. IEEE.

8. Yang, R., T.F. Tan, W. Lu, A.J. Thirunavukarasu, D.S.W. Ting, and N. Liu, *Large language models in health care: Development, applications, and challenges.* Health Care Science, 2023. **2**(4): p. 255-263.
9. Vavekanand, R., P. Karttunen, Y. Xu, S. Milani, and H. Li, *Large Language Models in Healthcare Decision Support: A Review.* 2024.
10. Islam, R. and O.M. Moushi, *Gpt-4o: The cutting-edge advancement in multimodal llm.* Authorea Preprints, 2024.
11. Jeblick, K., B. Schachtner, J. Dexl, A. Mittermeier, A.T. Stüber, J. Topalis, T. Weber, P. Wesp, B.O. Sabel, and J. Ricke, *ChatGPT makes medicine easy to swallow: an exploratory case study on simplified radiology reports.* European radiology, 2024. **34**(5): p. 2817-2825.
12. Casola, S. and A. Lavelli, *Summarization, simplification, and generation: The case of patents.* Expert Systems with Applications, 2022. **205**: p. 117627.
13. Koohi Habibi Dehkordi, M., Y. Perl, F.P. Deek, Z. He, V.K. Keloth, H. Liu, G. Elhanan, and A.J. Einstein, Improving Large Language Models' Summarization Accuracy by Adding Highlights to Discharge Notes: Comparative Evaluation. JMIR Medical Informatics, 2025. 13: p. e66476. doi: 10.2196/66476 . PMID: 40705416 . PMCID: 12332456.
14. Koohi Habibi Dehkordi, M., Y. Perl, F.P. Deek, and H. Liu, Fine-Tuning LLaMA2 for Summarizing Discharge Notes: Evaluating the Role of Highlighted Information. Big Data and Cognitive Computing, 2025. 10(1): p. 4.
15. Dashboard., F.A.E.R.S.F.P.; Available from: https://fis.fda.gov/sense/app/95239e26-e0be-42d9-a960-9a5f7f1c25ee/sheet/45beeb74-30ab-46be-8267-5756582633b4/state/analysis.
16. Donnelly, K., *SNOMED-CT: The advanced terminology and coding system for eHealth.* Stud Health Technol Inform, 2006. **121**: p. 279-90.
17. Rosenbloom, S.T., R.A. Miller, K.B. Johnson, P.L. Elkin, and S.H. Brown, *Interface terminologies: facilitating direct entry of clinical data into electronic health record systems.* Journal of the American medical informatics association, 2006. **13**(3): p. 277-288.
18. Duncker, E., J.A. Sheikh, and B. Fields. *From global terminology to local terminology: A review on cross-cultural interface design solutions*. in *Cross-Cultural Design. Methods, Practice, and Case Studies: 5th International Conference, CCD 2013, Held as Part of HCI International 2013, Las Vegas, NV, USA, July 21-26, 2013, Proceedings, Part I 5*. 2013. Springer.
19. Cimino, J.J., V.L. Patel, and A.W. Kushniruk, *Studying the human—computer—terminology interface.* Journal of the American Medical Informatics Association, 2001. **8**(2): p. 163-173.
20. Bakhshi-Raiez, F., L. Ahmadian, R. Cornet, E. de Jonge, and N. De Keizer, *Construction of an Interface Terminology on SNOMED CT.* Methods of Information in Medicine, 2010. **49**(04): p. 349-359.
21. Dehkordi, M.K.H., A.J. Einstein, S. Zhou, G. Elhanan, Y. Perl, V.K. Keloth, J. Geller, and H. Liu, *Using annotation for computerized support for fast skimming of cardiology electronic health record notes*, in *2023 IEEE International Conference on Bioinformatics and Biomedicine (BIBM)*. 2023, IEEE. p. 4043-4050.
22. Dehkordi, M.K.H., N.M. Kollapally, Y. Perl, J. Geller, F.P. Deek, H. Liu, V.K. Keloth, G. Elhanan, and A.J. Einstein. *Skimming of Electronic Health Records Highlighted by an Interface Terminology Curated with Machine Learning Mining*. in *BIOSTEC (2)*. 2024.
23. Johnson, A.E., T.J. Pollard, L. Shen, L.-w.H. Lehman, M. Feng, M. Ghassemi, B. Moody, P. Szolovits, L. Anthony Celi, and R.G. Mark, *MIMIC-III, a freely accessible critical care database.* Scientific data, 2016. **3**(1): p. 1-9.
24. github. *NLTK's list of english stopwords*. 2010; Available from: https://gist.github.com/sebleier/554280.
25. Association, A.S.-L.-H. *Common Medical Abbreviations*. Available from: https://www.asha.org/practice-portal/professional-issues/documentation-in-health-care/common-medical-abbreviations/.

26. Utah, U.o. *Cardiology Abbreviations and Diagnosis*. u.d; Available from: http://www.ped.med.utah.edu/pedsintranet/outpatient/triage/team_red/cardio_abbreviations_diagnosis.pdf.
27. Wikipedia. *List of medical abbreviations*. 2015; Available from: https://en.wikipedia.org/wiki/List_of_medical_abbreviations.
28. github. *Negex*. Available from: https://github.com/chapmanbe/negex/blob/master/genConText/trigger-neg.txt.
29. Heart.org. *Heart Medications*. u.d; Available from: https://www.heart.org/en/health-topics/heart-attack/treatment-of-a-heart-attack/cardiac-medications.
30. worldclasslearning. *1000 English Verbs Forms*. Available from: https://www.worldclasslearning.com/english/five-verb-forms.html#google_vignette.
31. Hardeniya, N., J. Perkins, D. Chopra, N. Joshi, and I. Mathur, *Natural language processing: python and NLTK*. 2016: Packt Publishing Ltd.
32. Daintith, J., *Kleene star*, in *A Dictionary of Computing*. 2008, Oxford University Press.
33. Almeida, F. and G. Xexéo, *Word embeddings: A survey.* arXiv preprint arXiv:1901.09069, 2019.
34. Alsentzer, E., J.R. Murphy, W. Boag, W.-H. Weng, D. Jin, T. Naumann, and M. McDermott, *Publicly available clinical BERT embeddings.* arXiv preprint arXiv:1904.03323, 2019.
35. Alimova, I. and E. Tutubalina, *Multiple features for clinical relation extraction: A machine learning approach.* Journal of biomedical informatics, 2020. **103**: p. 103382.
36. Dongare, A., R. Kharde, and A.D. Kachare, *Introduction to artificial neural network.* International Journal of Engineering and Innovative Technology (IJEIT), 2012. **2**(1): p. 189-194.
37. Abdi, H., D. Valentin, and B. Edelman, *Neural networks*. 1999: Sage.
38. Liashchynskyi, P. and P. Liashchynskyi, *Grid search, random search, genetic algorithm: a big comparison for NAS.* arXiv preprint arXiv:1912.06059, 2019.
39. Agarap, A.F., *Deep learning using rectified linear units (relu).* arXiv preprint arXiv:1803.08375, 2018.
40. Jais, I.K.M., A.R. Ismail, and S.Q. Nisa, *Adam optimization algorithm for wide and deep neural network.* Knowledge Engineering and Data Science, 2019. **2**(1): p. 41-46.
41. Fushiki, T., *Estimation of prediction error by using K-fold cross-validation.* Statistics and Computing, 2011. **21**: p. 137-146.
42. Srivastava, N., G. Hinton, A. Krizhevsky, I. Sutskever, and R. Salakhutdinov, *Dropout: a simple way to prevent neural networks from overfitting.* The journal of machine learning research, 2014. **15**(1): p. 1929-1958.
43. Bird, S., *NLTK: The Natural Language Toolkit.* ArXiv, 2004. **cs.CL/0205028**.
44. Kollapally, N.M., M.K.H. Dehkordi, Y. Perl, J. Geller, F.P. Deek, H. Liu, V.K. Keloth, G. Elhanan, A.J. Einstein, and S. Zhou. *Using clinical entity recognition for curating an interface terminology to aid fast skimming of EHRs*. in *2024 IEEE International Conference on Bioinformatics and Biomedicine (BIBM)*. 2024. IEEE.
45. White, J., Q. Fu, S. Hays, M. Sandborn, C. Olea, H. Gilbert, A. Elnashar, J. Spencer-Smith, and D.C. Schmidt, *A prompt pattern catalog to enhance prompt engineering with chatgpt.* arXiv preprint arXiv:2302.11382, 2023.
46. Giray, L., *Prompt engineering with ChatGPT: a guide for academic writers.* Annals of biomedical engineering, 2023. **51**(12): p. 2629-2633.
47. Haltaufderheide, J. and R. Ranisch, *The ethics of ChatGPT in medicine and healthcare: a systematic review on Large Language Models (LLMs).* NPJ digital medicine, 2024. **7**(1): p. 183.
48. Hadi, M.U., R. Qureshi, A. Shah, M. Irfan, A. Zafar, M.B. Shaikh, N. Akhtar, J. Wu, and S. Mirjalili, *A survey on large language models: Applications, challenges, limitations, and practical usage.* Authorea Preprints, 2023.
49. Karabacak, M. and K. Margetis, *Embracing large language models for medical applications: opportunities and challenges.* Cureus, 2023. **15**(5).

50. Miranda, M., E.S. Ruzzetti, A. Santilli, F.M. Zanzotto, S. Bratières, and E. Rodolà, *Preserving privacy in large language models: A survey on current threats and solutions.* arXiv preprint arXiv:2408.05212, 2024.
51. Yao, Y., J. Duan, K. Xu, Y. Cai, Z. Sun, and Y. Zhang, *A survey on large language model (llm) security and privacy: The good, the bad, and the ugly.* High-Confidence Computing, 2024: p. 100211.
52. Das, B.C., M.H. Amini, and Y. Wu, *Security and privacy challenges of large language models: A survey.* ACM Computing Surveys, 2024.
53. Zhou, B., D. Geißler, and P. Lukowicz, *Misinforming LLMs: vulnerabilities, challenges and opportunities.* arXiv preprint arXiv:2408.01168, 2024.
54. Perković, G., A. Drobnjak, and I. Botički. *Hallucinations in llms: Understanding and addressing challenges*. in *2024 47th MIPRO ICT and Electronics Convention (MIPRO)*. 2024. IEEE.
55. Dehkordi, M.K.H., J. Lu, Y. Perl, and F.P. Deek, *Enhancing Patient Comprehension of Discharge Notes with a Retrieval-Augmented LLM Approach*, in *2025 IEEE International Conference on Bioinformatics and Biomedicine (BIBM)*. 2025, IEEE.
56. Ranjan, R., S. Gupta, and S.N. Singh, *A comprehensive survey of bias in llms: Current landscape and future directions.* arXiv preprint arXiv:2409.16430, 2024.
57. Guo, Y., M. Guo, J. Su, Z. Yang, M. Zhu, H. Li, M. Qiu, and S.S. Liu, *Bias in large language models: Origin, evaluation, and mitigation.* arXiv preprint arXiv:2411.10915, 2024.
58. Ong, J.C.L., S.Y.-H. Chang, W. William, A.J. Butte, N.H. Shah, L.S.T. Chew, N. Liu, F. Doshi-Velez, W. Lu, and J. Savulescu, *Ethical and regulatory challenges of large language models in medicine.* The Lancet Digital Health, 2024. **6**(6): p. e428-e432.
59. Bedi, S., Y. Liu, L. Orr-Ewing, D. Dash, S. Koyejo, A. Callahan, J.A. Fries, M. Wornow, A. Swaminathan, and L.S. Lehmann, *A Systematic Review of Testing and Evaluation of Healthcare Applications of Large Language Models (LLMs).* medRxiv, 2024: p. 2024.04. 15.24305869.
60. Weiss, K., T.M. Khoshgoftaar, and D. Wang, *A survey of transfer learning.* Journal of Big data, 2016. **3**(1): p. 1-40.
61. Peng, Y., S. Yan, and Z. Lu. *Transfer Learning in Biomedical Natural Language Processing: An Evaluation of BERT and ELMo on Ten Benchmarking Datasets*. in *BioNLP@ACL*. 2019.
62. Sun, C. and Z. Yang. *Transfer Learning in Biomedical Named Entity Recognition: An Evaluation of BERT in the PharmaCoNER task*. in *Conference on Empirical Methods in Natural Language Processing*. 2019.
63. Giorgi, J. and G.D. Bader, *Transfer learning for biomedical named entity recognition with neural networks.* Bioinformatics, 2018. **34**: p. 4087 - 4094.
64. Dehkordi, M.K.H., Y. Perl, and F.P. Deek, *Optimizing Manual Review Using Machine Learning in Interface Terminology Curation for Automatic EHR Highlighting*, in *2025 IEEE International Conference on Bioinformatics and Biomedicine (BIBM)*. 2025, IEEE.